\documentclass[10pt,twocolumn,letterpaper]{article}

\usepackage{cvpr}
\usepackage{times}
\usepackage{epsfig}
\usepackage{graphicx}
\usepackage{amsmath}
\usepackage{amssymb}

\usepackage{echodefs} 

\usepackage{multirow}

\usepackage[pagebackref=true,breaklinks=true,letterpaper=true,colorlinks,bookmarks=false]{hyperref}

\cvprfinalcopy 

\def\cvprPaperID{17} 

\ifcvprfinal\fi

\begin{document}

\title{Improving the affordability of robustness training for DNNs}

\author{Sidharth Gupta\\
University of Illinois at Urbana-Champaign\\
{\tt\small gupta67@illinois.edu}
\and
Parijat Dube\\
IBM Research\\
{\tt\small pdube@us.ibm.com}
\and
Ashish Verma\\
IBM Research\\
{\tt\small ashish.verma1@ibm.com}
}

\maketitle

\begin{abstract}
Projected Gradient Descent (PGD) based adversarial training has become one of the most prominent methods for building robust deep neural network models. However, the computational complexity associated with this approach, due to the maximization of the loss function when finding adversaries, is a longstanding problem and may be prohibitive when using larger and more complex models. In this paper we show that the initial phase of adversarial training is redundant and can be replaced with natural training which significantly improves the computational efficiency. We demonstrate that this efficiency gain can be achieved without any loss in accuracy on natural and adversarial test samples. We support our argument with insights on the nature of the adversaries and their relative strength during the training process. We show that our proposed method can reduce the training time by a factor of up to 2.5 with comparable or better model test accuracy and generalization on various strengths of adversarial attacks.
\end{abstract}

\vspace{-2em}

\section{Introduction}

With the impressive performance of deep neural networks in multiple tasks, these models are being deployed in a variety of domains including entertainment, finance, healthcare, safety and security. However, a peculiar characteristic of these models is their extreme sensitivity to specially designed imperceptible small perturbations to the input data, called {\it Adversarial Samples}. For example, it is possible to make an otherwise highly accurate neural network classifier misclassify by adding a small imperceptible non-random perturbation to a test image \cite{szegedy2013intriguing}.

Out of the various approaches proposed to improve the robustness of deep neural network models, {\it Adversarial Training} is found to be most effective \cite{goodfellow2014explaining,szegedy2013intriguing,madry2017towards,wang2019convergence,cai2018curriculum,hendrycks2019using,duesterwald2019exploring}. In a typical adversarial training procedure, at each training iteration adversarial versions of the training dataset are generated and are then used to train the model to increase its robustness on such samples \cite{goodfellow2014explaining}. 

There are multiple methods to generate adversarial samples and these are known as the type of attack \cite{goodfellow2014explaining,carlini2017adversarial,sharma2017attacking,moosavi2017universal,madry2017towards}. The general principle behind most attacks is to identify data points in the input space which are imperceptibly close to the training data points but result in the highest loss function value. This results in the following general formulation
\begin{equation} \label{eq:attacks1}
    \max_{\tilde{\vx}} \calL(f_{\vtheta}({\tilde{\vx}}), y) \quad \text{s.t.} \quad d(\vx,\tilde{\vx}) \leq \epsilon 
\end{equation}
where $\calL$ is the loss function between the output of a classifier $f_{\vtheta}$ and the actual label $y$, $d$ is a distance metric between the original training sample $\vx$ and the corresponding adversarial sample $\tilde{\vx}$ and $\epsilon$ is a predetermined threshold. Different attack methods are designed by choosing different techniques to maximize Equation \eqref{eq:attacks1} and choosing different distance metrics. 

In this work, we use one of the most prominent adversarial training frameworks proposed by Madry et al. \cite{madry2017towards} which incorporates a min-max optimization of the overall objective function with respect to adversarial samples and model parameters. It has been shown to be one of the most effective training methods \cite{athalye2018obfuscated} and is also used as a state-of-the-art benchmark \cite{hendrycks2019using, wang2019convergence}. The models trained through this framework are shown to  be  robust against strong PGD attacks \cite{kurakin2016adversarial}, with MNIST \cite{lecun1998gradient} achieving about 90\% accuracy on adversarial samples. Although we have tested our approach with the framework proposed by Madry et al., our approach is generic and can be applied to other frameworks as well.

Adversarial training by Madry et al. \cite{madry2017towards} aims to solve the following min-max robust optimization problem,
\begin{equation} \label{eq:min-max}
    \min_{\vtheta} \rho(\vtheta); \quad \rho(\vtheta) = \mathbb{E}_{(\vx,y)\sim \calD} \left[\max_{\norm{\tilde{\vx} - \vx}_\infty \leq \epsilon} \calL(f_{\vtheta}(\tilde{\vx}), y)\right] ,
\end{equation}
where $f_{\vtheta}(.)$ is a deep neural network with parameters $\vtheta$ and $\tilde{\vx}$ is an $\epsilon$-ball $\ell_\infty$ adversarial sample of natural sample $\vx$ having class label $y$. $\tilde{\vx}$ is generated using Projected Gradient Descent (PGD) \cite{kurakin2016adversarial} as explained below in Equation \eqref{eq:pgd}. The true data comes from distribution $\calD$ and $\calL(\cdot)$ is the loss function. The maximization seeks to find an adversary which maximizes the loss and the minimization seeks to find the model parameters that minimizes the loss due to the adversary. The minimization is solved via standard neural network optimization techniques. The maximization is typically solved using Projected Gradient Descent (PGD),
\begin{equation} \label{eq:pgd}
    \vx^{t+1} = \Pi \left(\vx^t + \alpha\, \mathrm{sign}(\nabla_{\vx^t} \calL(f_{\vtheta}(\vx^t), y)) \right) ,
\end{equation}
where $\alpha$ is the step size and $\Pi(\cdot)$ projects the result of the gradient step into the $\epsilon$-ball around the original sample, $\vx$. It is a $T$-step PGD attack if $t+1 = T$. As the value of $T$ increases, the adversaries become stronger which results in a greater chance of misclassification by a trained model.

We are required to make $T$ forward and backward passes of the deep neural network to complete the iterative procedure in Equation~$\eqref{eq:pgd}$. This is a significant computational overhead to the training process and can be prohibitive for large models. This is especially relevant because models with larger capacity have been shown to be more robust \cite{madry2017towards}. Furthermore, some adversarial defense strategies involve training an ensemble of neural networks \cite{tramer2017ensemble,strauss2017ensemble,pang2019improving} and any per network training time reduction can scale quickly. 

Table \ref{table:natural_and_adv_times} confirms that the training time for adversarial training is significantly higher than natural training when classifying the CIFAR-10 image dataset \cite{krizhevsky2009learning} for two popular architectures. The same hardware, training dataset, training hyperparameters and total number of epochs were used to obtain all timings. Due to the popularity of adversarial training, we are motivated to improve its computational cost. Specifically, we question the need to always perform the expensive maximization, Equation~\eqref{eq:pgd}, in Equation~\eqref{eq:min-max}. We also wish to maintain the classification accuracy on both natural and adversarial test samples. Our method and insights are not restricted to this particular adersarial training framework and can be extended for other variants of adversarial training \cite{cai2018curriculum,carlini2017adversarial}. 

\begin{table}[tb]
\begin{center}
\begin{footnotesize}
\begin{sc}
\begin{tabular}{|c|c|c|}
\hline
\textbf{\begin{tabular}[c]{@{}c@{}}Model\\ architecture\end{tabular}} & \textbf{\begin{tabular}[c]{@{}c@{}}Natural\\ training\end{tabular}} & \textbf{\begin{tabular}[c]{@{}c@{}}Regular\\ adversarial\\ training\end{tabular}} \\ \hline
\textit{ResNet-50} & 1.1 hours & 6.8 hours \\ \hline
\textit{WideResNet-28x10} & 2.2 hours & 14.7 hours \\ \hline
\end{tabular}
\end{sc}
\end{footnotesize}
\end{center}
\caption{Natural and adversarial training times in hours for CIFAR-10 image dataset classification. All training runs are done for 155 epochs with two NVIDIA V100 GPUs, the same training hyperparameters and the standard CIFAR-10 training dataset split. Adversarial training is done using 10-step projected gradient descent adversaries with ball size $8/255$ and step size $2/255$.}
\label{table:natural_and_adv_times}
\vspace{-1em}
\end{table}


\paragraph{Our contribution.} We show that adversaries generated in the initial phase of adversarial training are treated more or less like natural samples by the final model. This means that they have minimal influence on the learned model and in fact may even hurt the final accuracy. As a consequence of this finding, we demonstrate that using natural, and not adversarial samples, for the initial phase of training gives comparable model test accuracy. We further show that this initial phase lasts for a specific number of epochs and therefore a fully converged naturally trained model cannot be taken as the initial phase. Importantly, our proposed training method significantly reduces the training time because the expensive maximization in \eqref{eq:min-max} is not required for a large fraction of the training epochs. 

\section{Related work} \label{sec:related_work}

The problem of overfitting and reduced generalization with regular adversarial training has been pointed out by several recent works including \cite{cai2018curriculum,wang2019convergence,schott2018towards,schmidt2018adversarially}. It has been attributed to training with strong adversarial samples from the beginning \cite{cai2018curriculum,wang2019convergence}. This has led to curriculum adversarial training \cite{cai2018curriculum} in which the strength of adversaries (as measured by the number of PGD steps in \cite{madry2017towards}) in a training batch is gradually increased as training progresses. Wang et al. further explained that the number of steps is not the right measure of adversary strength and they defined a new criterion by linking the strength of adversarial samples to the convergence of the inner maximization in Equation \ref{eq:min-max} \cite{wang2019convergence}. These works suggest that using lower quality adversaries during the initial phases of training helps improve accuracy. However, their focus is not computational efficiency and still requires adversaries to be computed in the initial phases of training. We also note that generalization of adversarially robust classifiers cannot be improved by algorithmic design alone as it is inherently tied to the complexity of the underlying data distribution \cite{schmidt2018adversarially}. 

The majority of research so far has focused on improving adversarial robustness through designing defenses and less consideration has been given to computational requirements and scalability, both of which are concerns with complex networks and big datasets. 
Shafahi et al.~\cite{shafahi2019adversarial} makes use of gradient information from model updates to reduce the overhead of generating adversarial samples and claims about 7x improvement in time for CIFAR-10 over regular adversarial training. However, multiple runs of their method may be required to select the parameters for their method which reduces the overall time saving. This is particularly important as they note an incorrect parameter drastically reduces natural accuracy. In a different line of work, ensembles of pretrained neural networks were used to reduce the computational cost of regular adversarial training in \cite{tramer2017ensemble} for ImageNet. The training was done by interspersing the adversarial samples generated using the trained network with adversarial samples generated from the ensemble, thereby creating a mixture of black-box and white-box attacks. 
Our approach can be used to efficiently develop pretrained networks for ensembles and can also be used to delay the generation of adversarial samples in the method proposed by Shafahi et al.
 
Adversarial training has also been accelerated by formulating it as a discrete-time differential game \cite{zhang2019you}. A recent work by Wong et al.~\cite{wong2020fast} claimed that adversarial training using FGSM with appropriate random initialization can be as effective as PGD thus making adversarial training much less costly. 
Though our work is focused on improving the efficiency of PGD based adversarial training, any method such as the one proposed by Wong et al. which requires adversarial samples to be generated during training can be made more efficient by using our method which delays the point at which adversarial samples are required.

The trade-offs between generalization and robustness are inherent \cite{tsipras2019robustness,su2018robustness}. Recently, there has been work suggesting that certain values of adversarial training hyperparameters may have unexpected benefits \cite{duesterwald2019exploring}. Duesterwald et al. used traditional approaches \cite{bergstra2011algorithms} to tune adversarial training hyperparameters to achieve different tradeoffs between robustness and accuracy. 
The appeal of our approach lies in its simplicity as its devoid of any (costly) hyperparameter optimization sub-steps. 

Switching from natural samples to adversarial samples can be interpreted as a transfer learning scenario. The current literature considers the scenario where the model is adversarially trained on both the source and target datasets to obtain better robustness \cite{hendrycks2019using}. Or where the model is adversarially trained on the source dataset and naturally trained on the target dataset \cite{shafahi2019adversarially}. In these approaches, while training on the target datasets may be fast because many epochs are not required, training the model on the source dataset is costly and we must efficiently make a model robust from the beginning.

\section{Delayed adversarial training} \label{sec:switching}

\subsection{Usefulness of adversaries from the initial epochs of regular adversarial training}

In the adversarial training framework natural training samples are replaced by their adversarial counterparts and used as training data from the start of training. Adversarial samples generated using Equation \eqref{eq:pgd} are dependent on the evolving model parameters which are initially randomly initialized. Generally at initialization, the model's parameters are relatively far from their final values. Therefore the adversarial samples generated in the initial training iterations are quite different from the adversaries that the model will face towards the end of training because they would not maximize the adversarial loss in Equation \eqref{eq:min-max} with the final model parameters. Hence, initial non-maximizing adversarial samples are weak adversaries for the final model and may not help improve robustness as was also noted by Wang et al. \cite{wang2019convergence}. Yet generating them adds computational overhead and they influence the model.

To investigate the usefulness of the initial adversarial samples we perform regular adversarial training on a model and test the final model with adversaries that are generated from the model's parameters at previous epochs. We use CIFAR-10 and the test images come from the dataset's standard train-test split which are never seen during training. We use the WideResNet-28x10 architecture \cite{zagoruyko2016wide} and adversarial test samples are generated using Equation~\eqref{eq:min-max} and \eqref{eq:pgd} with $T = 10, \epsilon = \frac{8}{255}$ and $\alpha = \frac{2}{255}$. This is a standard architecture and set of adversarial sample hyperparameters for CIFAR-10 in the adversarial training literature \cite{madry2017towards, hendrycks2019using}.

\begin{figure}[th]
    \centering
    \includegraphics[width=1.0\linewidth ]{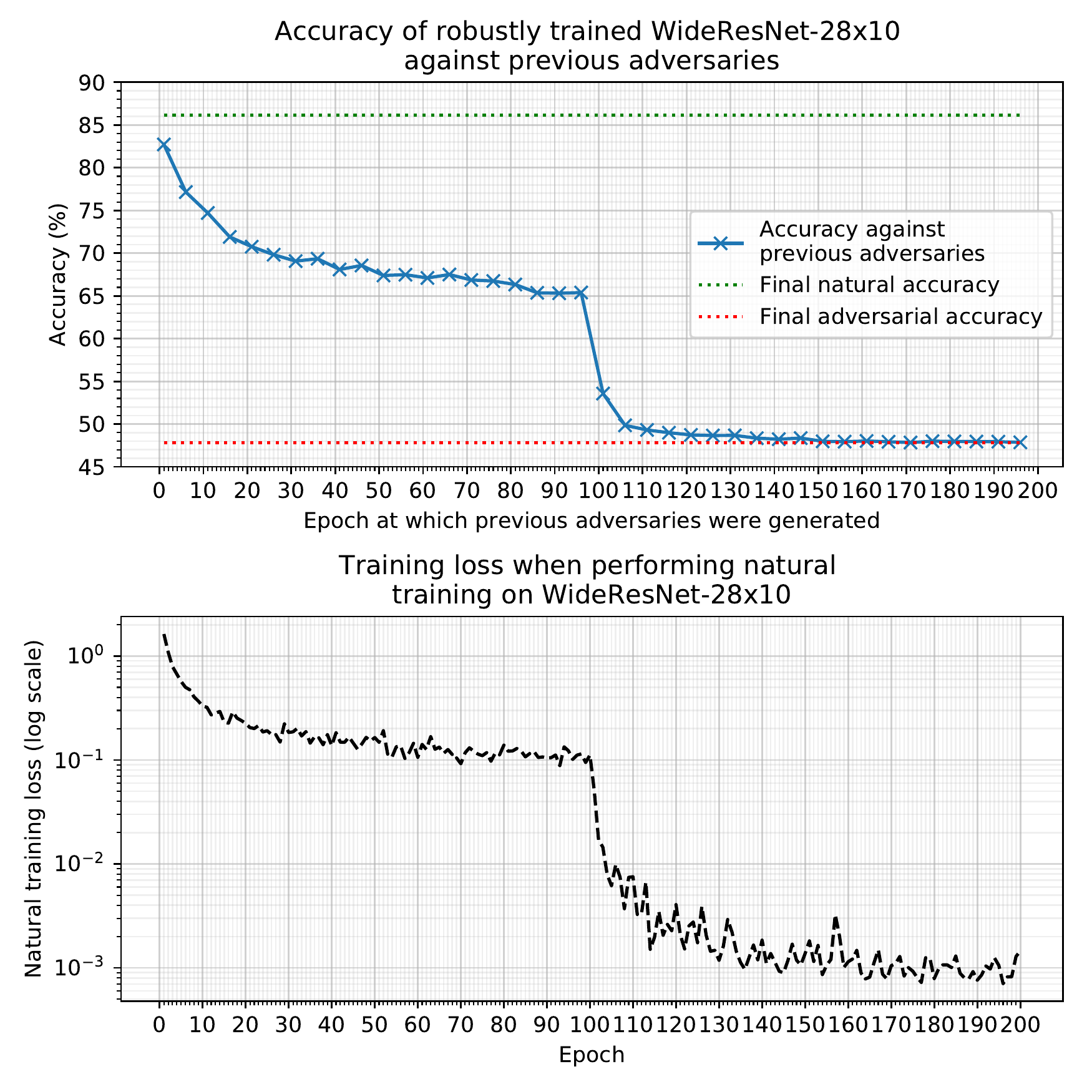}
    \caption{Top: Accuracy of a fully adversarially trained WideResNet-28x10 model when tested with adversaries that are generated using the model's parameters at previous epochs. The model is trained using the CIFAR-10 dataset. The green and red lines show the final model's test accuracy on natural and adversarial samples. Training and test samples are generated using \eqref{eq:min-max} and \eqref{eq:pgd} with $T = 10, \epsilon = \frac{8}{255}$ and $\alpha = \frac{2}{255}$. Stochastic gradient descent is used and the drop is due to a learning rate decrease; Bottom: The training loss (logarithmic scale on the right) when performing \emph{full} natural training on WideResNet-28x10 with CIFAR-10 samples.}
    \label{fig:previous_adversaries}
    \vspace{-1.0em}
\end{figure}

Figure \ref{fig:previous_adversaries} (Top) plots the classification accuracy when the final model is tested against adversaries from previous epochs. The green and red lines indicate the final natural and adversarial test accuracy of the models. We can see that adversarial samples from the initial epochs are treated more or less like natural samples by the final model. The adversaries become more potent as the model parameters start to approach their final value and the model starts to stabilize. The sharp increase in the adversary strength corresponds to the first learning rate drop as there is little change in model parameters from here on. We see that samples from the initial phase of training have limited impact on improving robustness.  In spite of this, the computationally expensive maximization \eqref{eq:pgd} is performed to generate these samples for training and these samples are allowed to influence the model parameters. Figure \ref{fig:resnet50_previous_adversaries} in the supplementary material shows that the same observation can be made for the ResNet-50.

\subsection{Delayed Adversarial Training (DAT): Initial training with natural samples}

As adversarial samples are computationally expensive to generate and not useful in the initial phase of training, we would like to replace them with natural training samples until the model reaches some form of stability. Natural samples for the initial training phase are an obvious choice as we also want the trained model to perform as well as possible on these samples.

Training on natural samples does not require the costly maximization in Equation \eqref{eq:pgd} and therefore significantly reduces the training time. Also the time taken to get the model to a state where adversarial samples are relevant reduces. Furthermore, the model is expected to correctly classify natural samples, but they are never seen in regular adversarial training. 

Now we investigate how to quantify the ``initial phase'' of training and determine a switch point from training on natural samples to training on adversarial samples.
In Figure~\ref{fig:previous_adversaries} (Bottom), we  plot the training loss evolution of full \textit{natural} training for the WideResNet-28x10 architecture with CIFAR-10. We see that the natural training loss flattens roughly at the same epoch at which the strength of the adversaries previously trained become reasonable in the top figure. This is due to the fact that at this point the model parameters have started to converge towards a local minimum. This motivates us to use the training loss on natural samples to determine when to switch to training on adversarial samples. We will use this motivation to propose an algorithm for a modified version of the adversarial training process later in this section.

\begin{figure*}
    \centering
    \includegraphics[width=1\linewidth ]{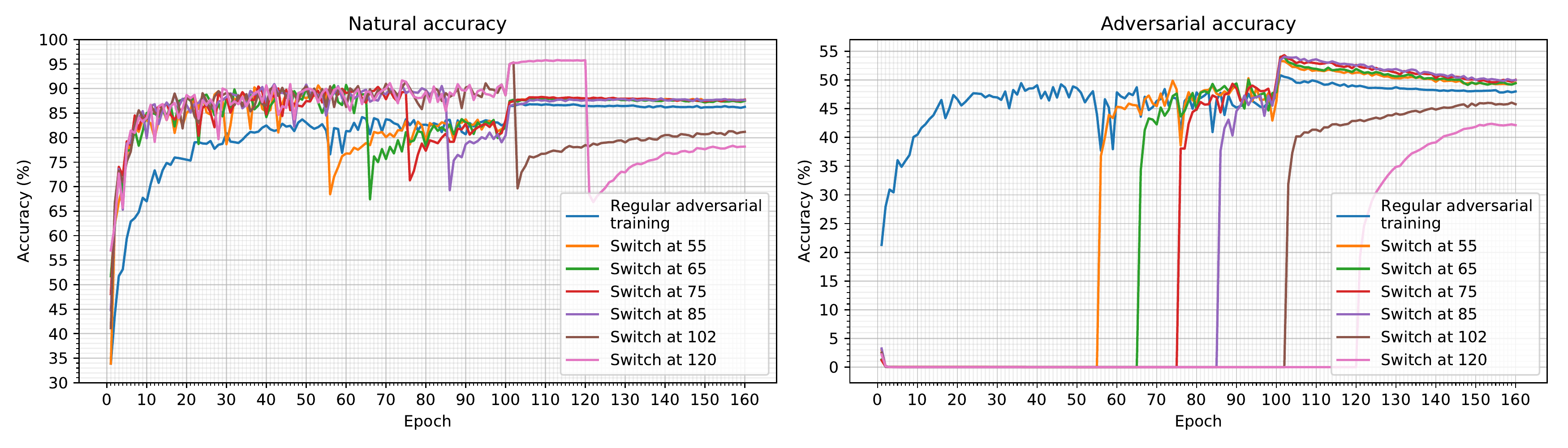}
    \caption{Natural and adversarial test accuracy during regular adversarial training and adversarial training with different switches. CIFAR-10 images are classified using the WideResNet-28x10. Adversarial samples with $T = 10, \epsilon = \frac{8}{255}$ and $\alpha = \frac{2}{255}$ are used. SGD learning rate drops are after epochs 100, 105 and 150. }
    \label{fig:wrn_triggers}
\end{figure*}

Before coming up with an exact algorithm for switching, we explore our proposition with observations on a set of experiments where we switched from natural training to adversarial training at various points. This is shown in Figure \ref{fig:wrn_triggers}. Here we trained the WideResNet-28x10 on the CIFAR-10 dataset. We performed the initial phase of training with natural samples and then switched to adversarial samples generated using PGD at various switching points. Training and test adversaries are generated using \eqref{eq:min-max} and \eqref{eq:pgd} with $T = 10, \epsilon = \frac{8}{255}$ and $\alpha = \frac{2}{255}$. The learning rate drops after epochs 100, 105 and 150. We show how the accuracy on the natural and adversarial test sets varies during regular adversarial training and adversarial training with different switching epochs.

Figure \ref{fig:wrn_triggers} provides evidence that performing the initial phase of training with natural samples is indeed a promising approach. We see that except for continuing the training with natural samples for too long (switching after the learning rate drops), we in fact get better performance than regular adversarial training. The training time is also reduced significantly as computing the expensive maximization in Equation~\eqref{eq:pgd} is not required until after the switch is made. Before switching, the adversarial accuracy is almost 0\% because the model is not robust and is undergoing natural training. Note, that the adversarial accuracy rises very quickly with all switching points as compared to the case of regular adversarial training where adversarial samples are used from the very beginning (the blue curve).
We observe similar curves for CIFAR-10 and CIFAR-100 classification with the ResNet-50 and ResNet-18 architectures in Section \ref{sec:additional_results} of the supplementary material.

Based on the above observation,  we first provide a basic algorithm (Algorithm~\ref{alg:switching_adv_training}) for our modification to the adversarial training method. We call it \textit{Delayed Adversarial Training (DAT)}. We introduce a new hyperparameter for delayed adversarial training, \textit{i.e.}, the epoch after which training data should be switched from natural samples to adversarial samples. We call this hyperparameter, the switching point, $S$. In Algorithm~\ref{alg:switching_adv_training}, $S$ must be chosen in advance. 
The value of this hyperprameter lies between $0$ and the total number of training epochs, $N$. For maximum computational saving, we would want to use natural samples for as long as possible and hence $S$ should be as close to $N$ as possible. However, as we have seen in Figure \ref{fig:wrn_triggers}, switching too late affects the final accuracy (both on natural samples and on adversarial samples). Specifically, we see this happening when the switch is after the learning rate drop, \textit{i.e.}, explicit actions to promote convergence because this will prevent the model from adapting to the newly introduced adversarial samples. Hence, a naive approach to naturally pretrain the model till full convergence and then switch to adversarial samples is not feasible.

\begin{algorithm}[tb]
\begin{small}
   \caption{Delayed Adversarial Training}
   \label{alg:switching_adv_training}
\begin{algorithmic} [1]
   \Require Switching hyperparameter $S$; $J$ training examples $\{(\vx_j, y_j) \}_{j=1}^J$; Number of epochs $N$; Optimizer and its parameters; PGD attack $\calA_{T, \epsilon, \alpha}$ with parameters $T, \epsilon, \alpha$ (number of steps, ball size, step size)
   \Ensure Model parameters, $\vtheta$
   \State Randomly initialize $\vtheta$
   \For{$i=0$ {\bfseries to} $N-1$}
   \ForAll {minibatch $(\vx_b, y_b)$}
   \If{$i > S$}
   \State Replace $\vx_b$ with $\vx_b \gets \calA_{T, \epsilon, \alpha}(\vtheta, \vx_b, y_b)$
   \EndIf \\
   // if condition replaces natural samples with adversarial samples after epoch $S$. $S = 0$ corresponds to regular adversarial training.
   \State Update $\vtheta$ with the optimizer using $(\vx_b, y_b)$
   \EndFor
   \EndFor
\end{algorithmic}
\end{small}
\end{algorithm}

Now we provide a method to automatically determine the value of the hyperparameter $S$. From Figure~\ref{fig:wrn_triggers}, a quick recipe for deciding $S$ is to choose an epoch after the test accuracy on natural samples flattens and before any learning rate drops. Hence, we use the evolution of training loss from Figure \ref{fig:previous_adversaries} to come-up with a dynamic strategy to switch from natural training to adversarial training. Specifically, we dynamically determine the value of $S$ as the epoch number where the training loss begins to converge. Convergence is determined if the training loss value at the current epoch is within $D$\% of the running average of training losses over the previous $W$ epochs. We explain Delayed Adversarial Training with this switching strategy in Algorithm \ref{alg:dynamic_switching_adv_training}.

\begin{algorithm}[tb]
\begin{small}
   \caption{Delayed Adversarial Training with Automated Switching}
   \label{alg:dynamic_switching_adv_training}
\begin{algorithmic} [1]
   \Require Length of training loss window to consider $W$; Training loss deviation $D$ from average window loss; $J$ training examples $\{(\vx_j, y_j) \}_{j=1}^J$; Number of epochs $N$; Optimizer and its parameters; PGD attack $\calA_{T, \epsilon, \alpha}$ with parameters $T, \epsilon, \alpha$ (number of steps, ball size, step size)
   \Ensure Model parameters $\vtheta$
   \State Randomly initialize $\vtheta$
   \State $S \gets N$ // initialize switch
   \For{$i=0$ {\bfseries to} $N-1$}
   \ForAll {minibatch $(\vx_b, y_b)$}
   \If{$i > S$}
   \State Replace $\vx_b$ with $\vx_b \gets \calA_{T, \epsilon, \alpha}(\vtheta, \vx_b, y_b)$
   \EndIf \\
   // if condition replaces natural samples with adversarial samples after epoch $S$.
   \State Update $\vtheta$ with optimizer and using $(\vx_b, y_b)$
   \EndFor
   \If {$S == N$}
   \State Calculate epoch's training loss $L_i$
   \If {$L_i$ is within $D$\% of the average of \{$L_{i-W}, \ldots, L_{i-2}, L_{i-1}\}$}
   \State $S \gets i$
   \EndIf
   \EndIf
   \EndFor
\end{algorithmic}
\end{small}
\end{algorithm}

It is easy to note that even though we have based our motivation and experimentation on the min-max approach \cite{madry2017towards}, our proposed modification to regular adversarial training can be easily extended to other frameworks which use different type of attacks in their training process.

\subsection{DAT helps generalization}

From Figure \ref{fig:wrn_triggers} we see that with delayed adversarial training,
the test accuracy is often better as compared to regular adversarial training. Also, there is a larger adversarial accuracy jump after the learning rate reduction at epoch 100 which is indicative of better generalization. This happens because in the case of delayed adversarial training, the model is not overfitting to adversaries of little relevance in the initial phase of training (see Figure \ref{fig:previous_adversaries}). Instead it is learning to classify natural samples which we always desire to be correctly classified. 

\begin{figure}[t]
    \centering
    \includegraphics[width=1.0\columnwidth]{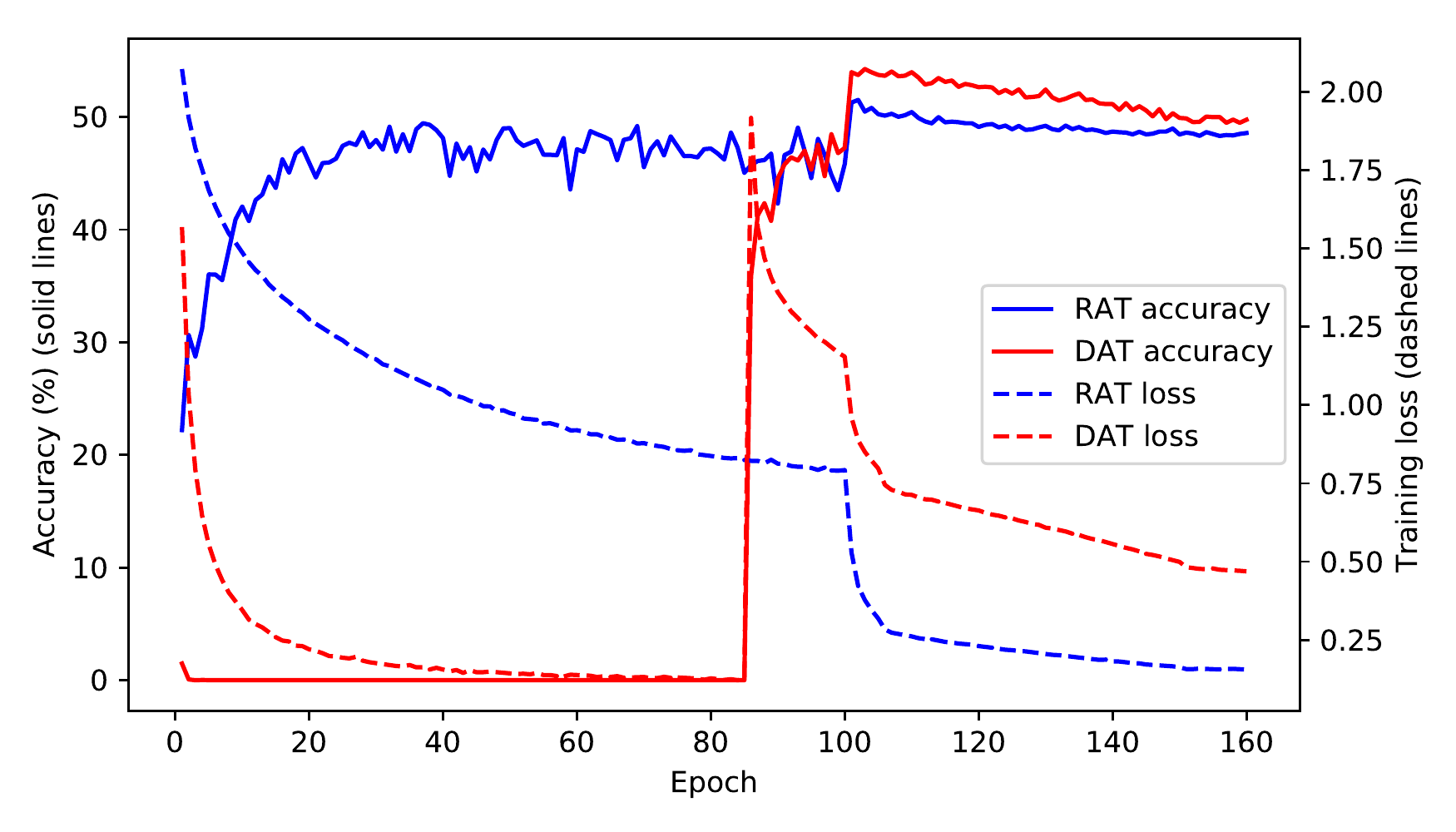}
    \caption{Adversarial test accuracy and training loss evolution for the WideResNet-28x10 for CIFAR-10. We show regular adversarial training (RAT) and delayed adversarial training (DAT). The left axis gives the accuracy and right axis gives the training loss. Solid lines are accuracy curves and dashed lines are training loss curves.}
    \label{fig:wrn_training_losses}
    \vspace{-0.75em}
\end{figure}

We can further confirm that our method helps generalization by looking at the training loss and adversarial test accuracy for regular adversarial training and delayed adversarial training. Figure \ref{fig:wrn_training_losses} shows how these values evolve during the course of training. In the case of delayed adversarial training (red dotted curve), the training loss is low until the switch, because the training has only been done on natural samples. This shoots up almost vertically at the time of switch to adversarial samples and then comes down gradually as the model get trained on adversarial samples. The learning rate drop at epoch 100 causes the training loss of both methods to drop significantly. As is evident from the figure, delayed adversarial training has higher training loss and higher test accuracy which indicates better generalization as compared to regular adversarial training. 

As mentioned in Section \ref{sec:related_work} Cai et al. \cite{cai2018curriculum} also demonstrated overfitting on adversarial samples with regular adversarial training. They show adversarial accuracy begins to drop during regular adversarial training. We additionally demonstrate this with ResNet-18 in supplementary material Section \ref{sec:resnet18}. We see that this drop is reduced with delayed adversarial training. Furthermore, Wang et al. \cite{wang2019convergence} also observed overfitting on adversarial samples when using strong adversarial samples in the initial phase of training. However, these works did not use this to make robustness training more efficient.

\section{Experiments} \label{sec:experiments}

In this section we evaluate our proposed method for efficiently training robust deep neural network models. 
We compare the training time and test accuracy of the models trained using our proposed method against regular adversarial training \cite{madry2017towards}.
We use PGD attacks which is one of the most powerful first-order $\ell_\infty$ attacks \cite{athalye2018obfuscated}. We use the standard CIFAR-10, CIFAR-100 and MNIST datasets in our evaluation.

For CIFAR-10, we use the WideResNet-28x10, ResNet-50 and ResNet-18 architectures with a batch size of 128. With CIFAR-100, we use ResNet-50 and ResNet-18 with a batch size of 128. For MNIST we use a model with two convolutional layers, which have 32 and 64 filters, followed by a fully connected layer of size 1024. Each convolutional layer is followed by a ReLU and $2 \times 2$ max pooling. The batch size for MNIST experiments is 50.

SGD with momentum 0.9 and weight decay $2 \times 10^{-4}$ is used to train the WideResNet-28x10 and all the ResNets for CIFAR. The initial learning rate for the WideResNet-28x10 is 0.1 and it is reduced by a factor of ten after epochs 100, 105 and 150 and the model is trained for a total of 155 epochs. The ResNets also have an initial learning rate of 0.1 which is reduced by a factor of ten after epochs 100 and 150 and the model is trained for a total of 155 epochs. Adam with an initial learning rate of $1 \times 10^{-4}$ is used for MNIST. The Adam optimizer state and learning rate is reinitialized after switching. MNIST training is for 80 epochs. $W$ and $D$ parameters from Algorithm \ref{alg:dynamic_switching_adv_training} are empirically tuned.

We parameterized PGD adversaries as $\{T, \epsilon, \alpha\}$ where $T$ is the number of PGD steps, $\epsilon$ is the maximum $\ell_\infty$ perturbation and $\alpha$ is the gradient ascent step size (see Equations \eqref{eq:min-max} and \eqref{eq:pgd}). For CIFAR-10 and CIFAR-100 all adversarial samples used during training are of strength $\left\{10, \frac{8}{255}, \frac{2}{255}\right\}$. MNIST adversarial samples for training are generated with $\{40, 0.3, 0.01\}$. These are standard strengths in the literature \cite{madry2017towards, hendrycks2019using}. All PGD adversaries are constructed by adding an initial random perturbation, $\vdelta$, where $\norm{\vdelta}_\infty \leq \epsilon$.

CIFAR experiments are performed on a system with two NVIDIA V100 GPUs, 42 CPU cores and 48GB memory. For MNIST experiments we use a system with one NVIDIA V100 GPU, 26 CPU cores and 24GB memory.

As ultimately the time taken to reach a given accuracy is what matters and not the total number of epochs, we also have an early stopping criteria for the CIFAR-10 models to regularize and avoid training the models for too long without much gain in accuracy. In this case, we  stop training after the first learning rate drop once the test loss on the natural samples of the current epoch is within 5\% of the average test loss of the previous five epochs. Furthermore early stopping helps prevent overfitting to adversarial samples.


\subsection{Training times}

Table \ref{table:all_times} shows the training times for all the models and datasets. We compare the time taken for regular adversarial training and our method. The timings for the early stopping of CIFAR-10 models are also shown. Note that MNIST experiments are done on a less powerful system than the CIFAR experiments.

\begin{table*}
\begin{center}
\begin{footnotesize}
\begin{sc}
\begin{tabular}{|c|c|c|c|c|c|}
\hline
\multicolumn{2}{|l|}{\multirow{3}{*}{}} & \multicolumn{4}{c|}{\multirow{2}{*}{\textbf{CIFAR-10}}} \\
\multicolumn{2}{|l|}{} & \multicolumn{4}{c|}{} \\ \cline{3-6} 
\multicolumn{2}{|l|}{} & \textit{Training time} & \textit{Time saved} &  \textit{$T=10$, $\epsilon = 8/255$} & \textit{Natural accuracy}\\ \hline
\multirow{4}{*}{\textit{WideResNet-28x10}} & \textit{RAT} & 14.7 hours & \multirow{2}{*}{46.9\%} & 48.5\% & 86.8\% \\ \cline{2-3} \cline{5-6} 
 & \textit{DAT} & 7.8 hours &  & 49.7\% & 87.9\% \\ \cline{2-6} 
 & \textit{RAT early stop} & 10.9 hours & \multirow{2}{*}{62.4\%} & 49.2\% & 87.1\% \\ \cline{2-3} \cline{5-6} 
 & \textit{DAT early stop} & 4.1 hours &  & 53.6\% & 87.9\%  \\ \hline
\multirow{4}{*}{\textit{ResNet-50}} & \textit{RAT} & 6.8 hours & \multirow{2}{*}{45.6\%} & 41.0\% & 75.2\% \\ \cline{2-3} \cline{5-6} 
 & \textit{DAT} & 3.7 hours &  & 41.1\% & 74.4\% \\ \cline{2-6} 
 & \textit{RAT early stop} & 5.0 hours & \multirow{2}{*}{64.0\%} & 42.3\% & 73.2\% \\ \cline{2-3} \cline{5-6} 
 & \textit{DAT early stop} & 1.8 hours &  & 41.6\% & 72.2\%  \\ \hline
\multirow{4}{*}{\textit{ResNet-18}} & \textit{RAT} & 2.5 hours & \multirow{2}{*}{36.0\%} & 37.0\% & 73.8\% \\ \cline{2-3} \cline{5-6} 
 & \textit{DAT} & 1.6 hours &  & 40.4\% & 72.8\% \\ \cline{2-6} 
 & \textit{RAT early stop} & 1.9 hours & \multirow{2}{*}{52.6\%} & 40.6\% & 71.0\% \\ \cline{2-3} \cline{5-6} 
 & \textit{DAT early stop} & 0.9 hours &  & 41.1\% & 69.9\% \\ \hline
\multicolumn{2}{|c|}{\multirow{3}{*}{\textit{}}} & \multicolumn{4}{c|}{\multirow{2}{*}{\textbf{CIFAR-100}}} \\
\multicolumn{2}{|c|}{} & \multicolumn{4}{c|}{} \\ \cline{3-6} 
\multicolumn{2}{|c|}{} & \textit{Training time} & \textit{Time saved} &  \textit{$T=10$, $\epsilon = 8/255$} & \textit{Natural accuracy} \\ \hline
\multirow{2}{*}{\textit{ResNet-50}} & \textit{RAT} & 6.9 hours & \multirow{2}{*}{42.0\%} & 15.2\% & 44.2\% \\ \cline{2-3} \cline{5-6} 
 & \textit{DAT} & 4.0 hours &  & 15.2\% & 46.6\% \\ \hline
\multirow{2}{*}{\textit{Resnet-18}} & \textit{RAT} & 2.6 hours & \multirow{2}{*}{46.2\%} & 14.2\% & 44.7\% \\ \cline{2-3} \cline{5-6} 
 & \textit{DAT} & 1.4 hours &  & 14.2\% & 46.7\% \\ \hline
\multicolumn{2}{|c|}{\multirow{3}{*}{}} & \multicolumn{4}{c|}{\multirow{2}{*}{\textbf{MNIST}}} \\
\multicolumn{2}{|c|}{} & \multicolumn{4}{c|}{} \\ \cline{3-6} 
\multicolumn{2}{|c|}{} & \textit{Training time} & \textit{Time saved} & \textit{$T=40$, $\epsilon = 0.3$} & \textit{Natural accuracy} \\ \hline
\multirow{2}{*}{\textit{Two-layer CNN}} & \textit{RAT} & 2.2 hours & \multirow{2}{*}{13.6 \%} & 91.4\% & 98.2\% \\ \cline{2-3} \cline{5-6} 
 & \textit{DAT} & 1.9 hours &  & 91.9\% & 98.2\% \\ \hline
\end{tabular}
\end{sc}
\end{footnotesize}
\end{center}
\caption{Training time and test accuracy with natural samples and adversaries of same strength as training adversaries for Regular adversarial training (RAT) and Delayed Adversarial Training (DAT). MNIST experiments are done on a less powerful system.}
\label{table:all_times}
\vspace{-1em}
\end{table*}

The training times are significantly reduced using our method with often better accuracy. The importance of our method is particularly felt when training large models such as the WideResNet-28x10 on the more complex CIFAR-10 dataset. In this case we get a 46.9\% reduction in training time. With regularization via early stopping, we get 62.4\% savings in training time. On all other dataset and models also we get significant reduction in training time with accuracy very close to that of regular adversarial training. In fact, our robustness accuracy is mostly higher as compared to that of regular adversarial training due to better generalization on adversarial samples as discussed in Section \ref{sec:switching}.

\subsection{Generalization to other attacks} \label{sec:generalization}

\begin{figure}
    \centering
    \includegraphics[width=1\linewidth]{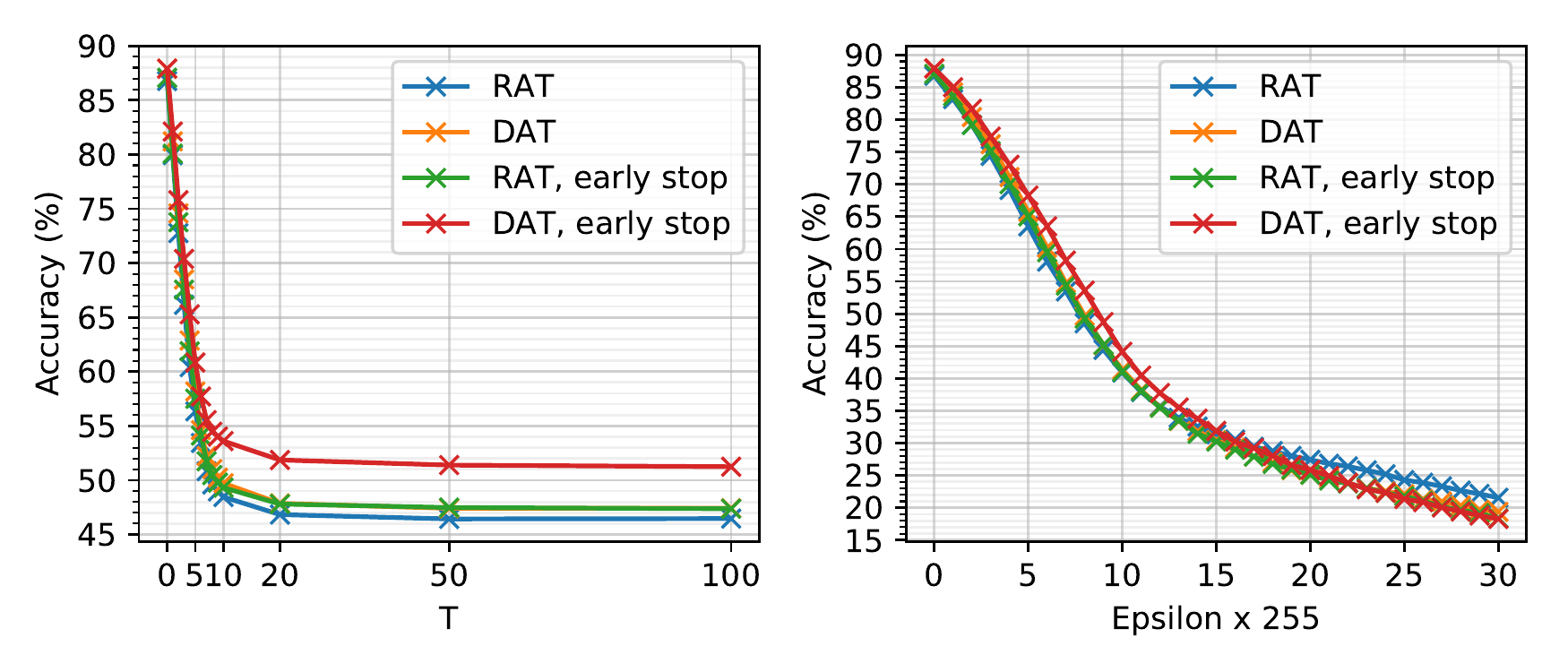}
    \caption{Accuracy of fully trained WideResNet-28x10 with CIFAR-10 when tested with attacks of different strength. Adversaries used during training were of strength $\left\{10, \frac{8}{255}, \frac{2}{255}\right\}$.}
    \label{fig:wrn_resistance}
    \vspace{-0.75em}
\end{figure}

\begin{figure}
    \centering
    \includegraphics[width=1\linewidth]{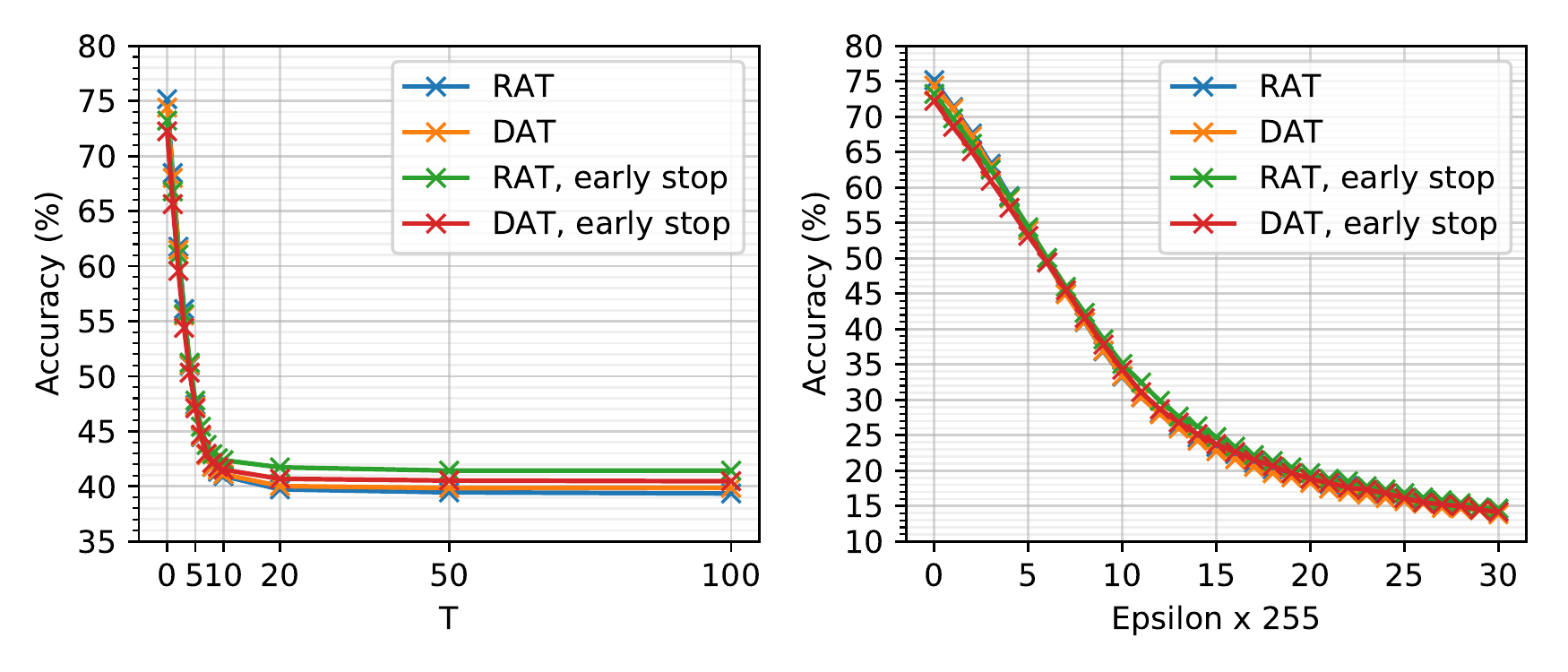}
    \caption{Accuracy of fully trained ResNet-50 with CIFAR-10 when tested with attacks of different strength. Adversaries used during training were of strength $\left\{10, \frac{8}{255}, \frac{2}{255}\right\}$.}
    \label{fig:resnet50_resistance}
    \vspace{-0.75em}
\end{figure}

\begin{figure}
    \centering
    \includegraphics[width=1\linewidth ]{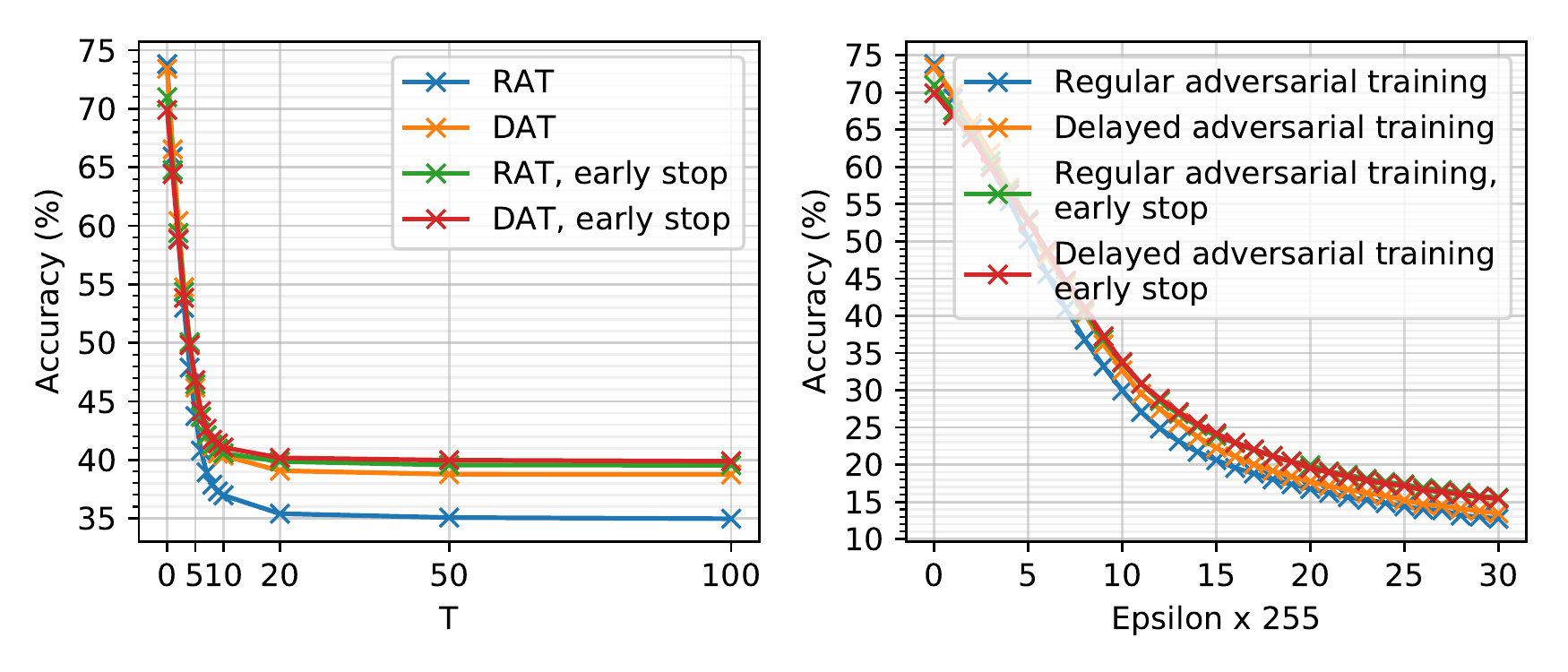}
    \caption{Accuracy of fully trained ResNet-18 with CIFAR-10 when tested with attacks of different strength. Adversaries used during training were of strength $\left\{10, \frac{8}{255}, \frac{2}{255}\right\}$.}
    \label{fig:resnet18_resistance}
    \vspace{-0.75em}
\end{figure}

\begin{figure}
    \centering
    \includegraphics[width=1\linewidth ]{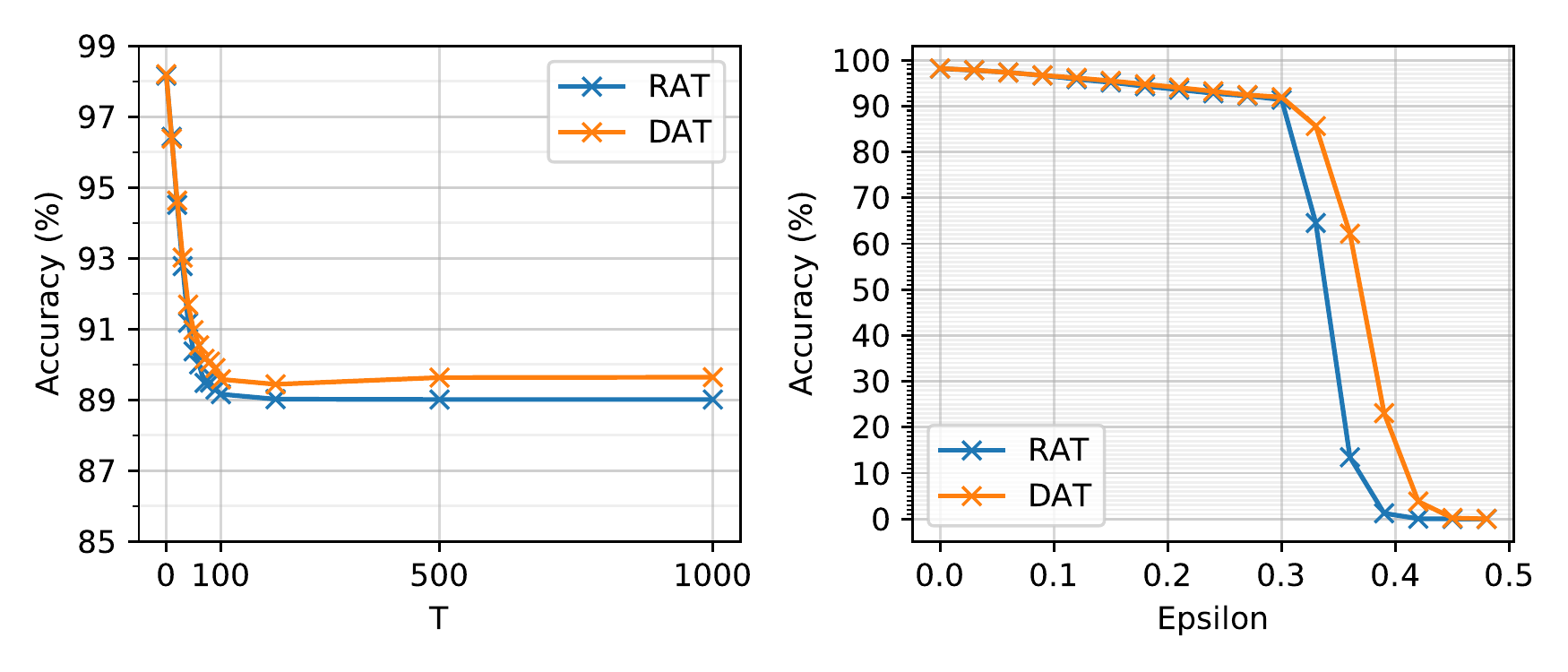}
    \caption{Accuracy of fully trained two-layer CNN with MNIST when tested with attacks of different strength. Adversaries used during training were of strength $\left\{40, 0.3, 0.01\right\}$.}
    \label{fig:mnist_resistance}
    \vspace{-0.75em}
\end{figure}

We evaluate the robustness of our models against attacks of strengths which they were not trained to resist. We vary the number of PGD steps and the $\epsilon$-ball size of the test attacks. Figures \ref{fig:wrn_resistance}, \ref{fig:resnet50_resistance} and \ref{fig:resnet18_resistance} show the performance with regular adversarial training and our method for the CIFAR-10 models (Figure \ref{fig:resnet18_cf100_resistance} in supplementary material Section \ref{sec:resnet18} shows the robustness for CIFAR-100 with ResNet-18). Figure \ref{fig:mnist_resistance} shows it for MNIST. The models trained using our method are comparably robust against a wide variety of attacks and follow the same pattern as seen in Madry et al. \cite{madry2017towards}. As expected~\cite{athalye2018obfuscated} a larger  number of PGD iterations and $\epsilon$-ball size decreases accuracy. Table~\ref{table:all_generalization} in the supplementary material summarizes the differences between regular adversarial training and our method for typical benchmark adversaries.

We note that increasing the model capacity improves robustness for all datasets which is consistent with observations made by Madry et al. Also consistent is the sharp fall in MNIST robustness when tested with $\epsilon$ larger than what the model was trained for ($\epsilon=0.3$). Our method delays the collapse in robustness as we can see from Figure~\ref{fig:mnist_resistance}. 

\paragraph{Black-box attacks}

We check the performance of models trained using our method under black-box attacks from independently trained copies of the network. We test with CIFAR-10 and the WideResNet-28x10 with adversaries of strength $\left\{10, \frac{8}{255}, \frac{2}{255}\right\}$. From Table \ref{table:all_times} the white-box accuracy using delayed adversarial training is $49.7\%$.

Independent copies of the network were trained with 1) Natural training; 2) Regular adversarial training (RAT); 3) Delayed adversarial training (DAT). Table \ref{table:black_box} shows that black-box accuracy is higher than white-box accuracy as expected \cite{athalye2018obfuscated} and reports accuracy values that are consistent with prior work \cite{madry2017towards}.

\begin{table}[h]
\begin{center}
\begin{footnotesize}
\begin{sc}
\begin{tabular}{|c|c|c|c|c|}
\hline
\textit{Source} & \textbf{White-box} & \textbf{Natural} & \textbf{RAT} & \textbf{DAT} \\ \hline
\textit{Accuracy} & 49.7\% & 86.7\% & 69.4\% & 65.7\% \\ \hline
\end{tabular}
\end{sc}
\end{footnotesize}
\end{center}
\caption{CIFAR-10 Black-box robustness against attacks from independent copies of the WideResNet-28x10. }
\label{table:black_box}
\vspace{-0.5em}
\end{table}

\subsection{Performance with modified training schedules}

\begin{figure}[ht]
    \centering
    \includegraphics[width=1\linewidth ]{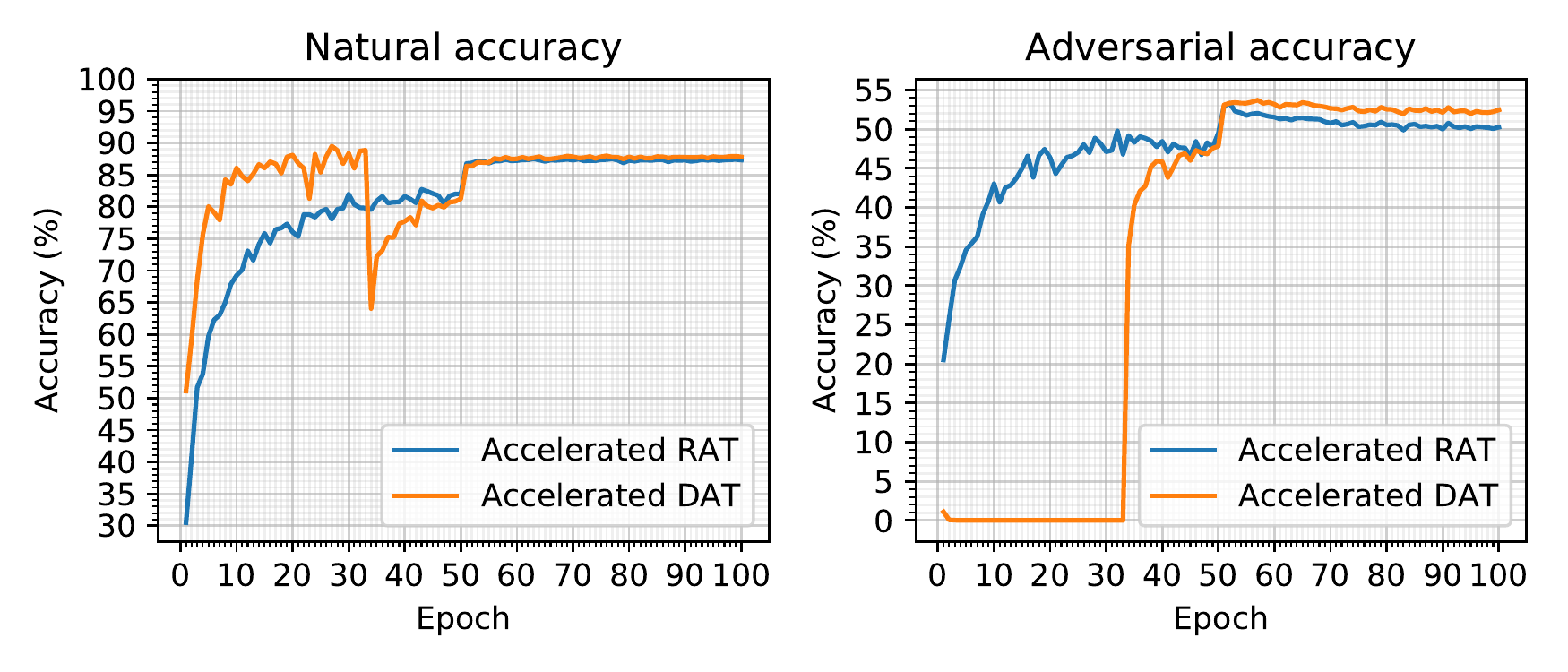}
    \caption{Natural and adversarial test accuracy during regular adversarial training and delayed adversarial training when the training process is accelerated. CIFAR-10 images are classified using the WideResNet-28x10. Adversarial samples with $T = 10, \epsilon = \frac{8}{255}$ and $\alpha = \frac{2}{255}$ are used.}
    \label{fig:wrn_half_triggers}
    \vspace{-1.5em}
\end{figure}

Until now we used similar training and learning rate schedules across models to fairly compare against regular adversarial training accuracy values reported in literature. Next we show that our results are independent of the schedule. We reduce the adversarial training time by accelerating the learning rate drops and finishing the training process in a relatively smaller number of epochs. The total number of epochs, learning rate drop epochs and the switching hyperparameter $S$, are approximately halved. Figure \ref{fig:wrn_half_triggers} shows the natural and adversarial test accuracy evolution during training for this accelerated training procedure for CIFAR-10 with the WideResNet-28x10 model. The measured training times for the regular adversarial training and our proposed delayed adversarial training for this experiments are 9.5 hours and 6.8 hours respectively---a 28\% saving. Furthermore, we see that this training schedule gives an almost 2\% improvement in adversarial accuracy over regular adversarial training.

\begin{figure}[ht]
    \centering
    \includegraphics[width=1\linewidth ]{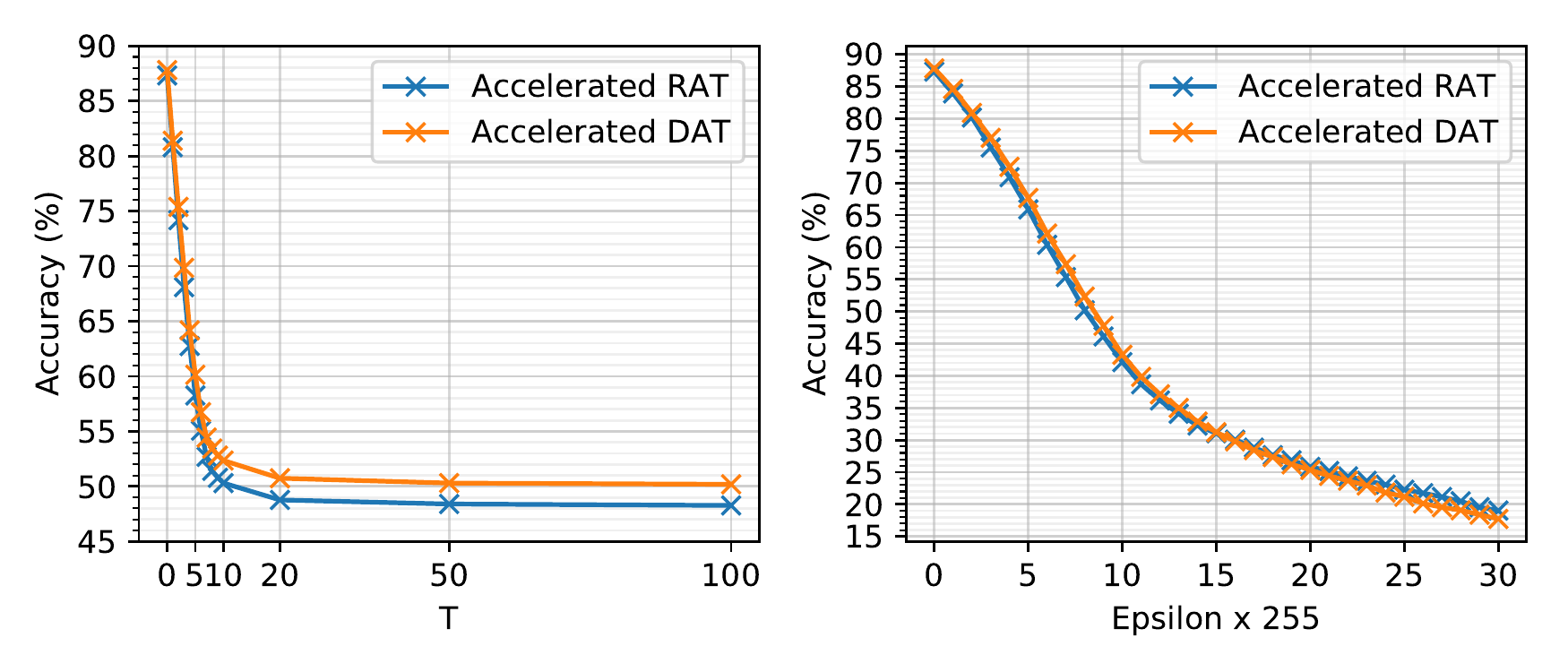}
    \caption{Accuracy of fully trained WideResNet-28x10 with CIFAR-10 when training is accelerated and tested with attacks of different strength. Adversaries used during training were of strength $\left\{10, \frac{8}{255}, \frac{2}{255}\right\}$.}
    \label{fig:wrn_half_resistance}
    \vspace{-1em}
\end{figure}

Figure \ref{fig:wrn_half_resistance} shows the performance of the models trained with accelerated training against attacks of different strengths. There is a slight reduction in overall performance of the accelerated training models at higher values of $\epsilon$. However, when only comparing accelerated models, the performance of the models trained with our proposed method is comparable to the model trained with regular adversarial training.

\section{Conclusion}
In this paper we analyzed the computational efficiency of regular adversarial training to efficiently improve the robustness of deep neural networks. We presented insights about the usefulness of training with adversarial samples in the initial and later phases of regular adversarial training and proposed a modified version of the training framework which significantly improves its computational requirement. We further show through various experiments that the neural network models trained through the proposed framework are as accurate or better as the ones trained through the earlier framework and generalize quite well to adversarial attacks of various strengths. 

In the future, we plan to adapt our general method to other adversarial training frameworks. Furthermore, although our goal is to improve training time, we get better accuracy in many cases. Future work aims to explore this further to develop a methodology which optimizes accuracy and computational efficiency in a more integrated way.

\clearpage

\clearpage

\section*{Supplementary material}

\appendix






\section{Summary of generalization to other attacks}

Table \ref{table:all_generalization} summarizes the results from Section \ref{sec:generalization} on generalization to attacks of strengths which models were not trained to resist. Selected values for typical benchmark adversaries from Figures \ref{fig:wrn_resistance}, \ref{fig:resnet50_resistance}, \ref{fig:resnet18_resistance}, \ref{fig:resnet18_cf100_resistance} and \ref{fig:mnist_resistance} are summarized in the table.

\begin{table*}[t]
\begin{center}
\begin{footnotesize}
\begin{sc}
\begin{tabular}{|c|c|c|c|c|c|}
\hline
\multicolumn{2}{|l|}{\multirow{3}{*}{}} & \multicolumn{4}{c|}{\multirow{2}{*}{\textbf{CIFAR-10}}} \\
\multicolumn{2}{|l|}{} & \multicolumn{4}{c|}{} \\ \cline{3-6} 
\multicolumn{2}{|l|}{} & \textit{\begin{tabular}[c]{@{}c@{}}$T=20$,\\ $\epsilon = 8/255$\end{tabular}} & \textit{\begin{tabular}[c]{@{}c@{}}$T=100$, \\ $\epsilon = 8/255$\end{tabular}} & \textit{\begin{tabular}[c]{@{}c@{}}$T=10$, \\ $\epsilon = 4/255$\end{tabular}} & \textit{\begin{tabular}[c]{@{}c@{}}$T=10$, \\ $\epsilon = 12/255$\end{tabular}} \\ \hline
\multirow{4}{*}{\textit{WideResNet-28x10}} & \textit{RAT} & 46.8\% & 46.5\% & 69.1\% & 35.7\% \\ \cline{2-6} 
 & \textit{DAT} & 47.9\% & 47.4\% & 71.1\% & 35.5\% \\ \cline{2-6} 
 & \textit{RAT early stop} & 47.8\% & 47.4\% & 70.0\% & 35.5\% \\ \cline{2-6} 
 & \textit{DAT early stop} & 51.9\% & 51.2\% & 73.0\% & 37.7\% \\ \hline
\multirow{4}{*}{\textit{ResNet-50}} & \textit{RAT} & 39.7\% & 39.4\% & 58.8\% & 28.1\% \\ \cline{2-6} 
 & \textit{DAT} & 40.0\% & 39.9\% & 58.5\% & 28.0\% \\ \cline{2-6} 
 & \textit{RAT early stop} & 41.7\% & 41.4\% & 58.6\% & 29.9\% \\ \cline{2-6} 
 & \textit{DAT early stop} & 40.7\% & 40.5\% & 57.1\% & 28.7\% \\ \hline
\multirow{4}{*}{\textit{ResNet-18}} & \textit{RAT} & 35.4\% & 35.0\% & 55.4\% & 24.9\% \\ \cline{2-6} 
 & \textit{DAT} & 39.1\% & 38.8\% & 57.3\% & 27.4\% \\ \cline{2-6} 
 & \textit{RAT early stop} & 39.9\% & 39.6\% & 56.9\% & 28.6\% \\ \cline{2-6} 
 & \textit{DAT early stop} & 40.2\% & 39.9\% & 56.4\% & 28.9\% \\ \hline
\multicolumn{2}{|c|}{\multirow{3}{*}{\textit{}}} & \multicolumn{4}{c|}{\multirow{2}{*}{\textbf{CIFAR-100}}} \\
\multicolumn{2}{|c|}{} & \multicolumn{4}{c|}{} \\ \cline{3-6} 
\multicolumn{2}{|c|}{} & \textit{\begin{tabular}[c]{@{}c@{}}$T=20$,\\  $\epsilon = 8/255$\end{tabular}} & \textit{\begin{tabular}[c]{@{}c@{}}$T=100$, \\ $\epsilon = 8/255$\end{tabular}} & \textit{\begin{tabular}[c]{@{}c@{}}$T=10$, \\ $\epsilon = 4/255$\end{tabular}} & \textit{\begin{tabular}[c]{@{}c@{}}$T=10$, \\ $\epsilon = 12/255$\end{tabular}} \\ \hline
\multirow{2}{*}{\textit{ResNet-50}} & \textit{RAT} & 14.6\% & 14.2\% & 25.7\% & 9.8\% \\ \cline{2-6} 
 & \textit{DAT} & 14.5\% & 14.3\% & 26.4\% & 9.5\% \\ \hline
\multirow{2}{*}{\textit{ResNet-18}} & \textit{RAT} & 13.6\% & 13.1\% & 25.3\% & 8.7\% \\ \cline{2-6} 
 & \textit{DAT} & 13.5\% & 13.1\% & 26.4\% & 8.7\% \\ \hline
\multicolumn{2}{|c|}{\multirow{3}{*}{}} & \multicolumn{4}{c|}{\multirow{2}{*}{\textbf{MNIST}}} \\
\multicolumn{2}{|c|}{} & \multicolumn{4}{c|}{} \\ \cline{3-6} 
\multicolumn{2}{|c|}{} & \textit{\begin{tabular}[c]{@{}c@{}}$T=100$, \\ $\epsilon = 0.3$\end{tabular}} & \textit{\begin{tabular}[c]{@{}c@{}}$T=1000$, \\ $\epsilon = 0.3$\end{tabular}} & \textit{\begin{tabular}[c]{@{}c@{}}$T=40$, \\ $\epsilon = 0.33$\end{tabular}} & \textit{\begin{tabular}[c]{@{}c@{}}$T=40$, \\ $\epsilon = 0.36$\end{tabular}} \\ \hline
\multirow{2}{*}{\textit{Two-layer CNN}} & \textit{RAT} & 89.2\% & 89.0\% & 64.5\% & 13.4\% \\ \cline{2-6} 
 & \textit{DAT} & 89.6\% & 89.6\% & 85.6\% & 62.2\% \\ \hline
\end{tabular}
\end{sc}
\end{footnotesize}
\end{center}
\caption{Robustness of models against adversaries with strengths that they were not trained to be robust against when using Regular Adversarial Training (RAT) and Delayed Adversarial Training (DAT).}
\label{table:all_generalization}
\end{table*}

\section{Additional results} \label{sec:additional_results}

\subsection{Additional ResNet-50 results}

\begin{figure}[h]
    \centering
    \includegraphics[width=1\columnwidth]{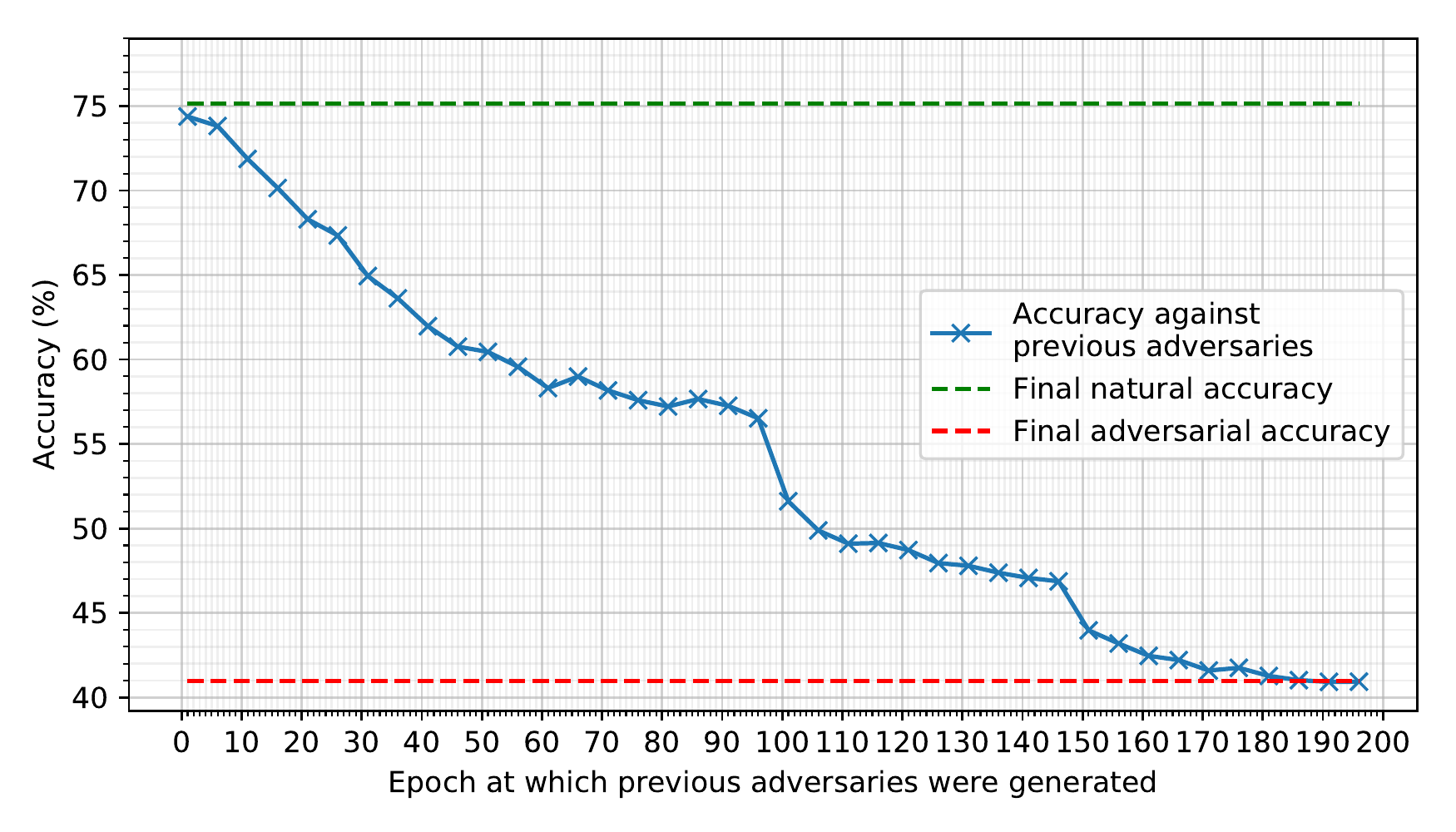}
    \caption{Accuracy of a fully adversarially trained ResNet-50 model when tested with adversaries that are generated using the model's parameters at previous epochs. The model is trained using the CIFAR-10 dataset. The green and red lines show the final model's test accuracy on natural and adversarial samples. CIFAR-10 training and test samples are generated using \eqref{eq:min-max} and \eqref{eq:pgd} with $T = 10, \epsilon = \frac{8}{255}$ and $\alpha = \frac{2}{255}$. Stochastic gradient descent is used and the drops are due to learning rate decreases.}
    \label{fig:resnet50_previous_adversaries}
\end{figure}

Figure \ref{fig:resnet50_previous_adversaries} shows the test accuracy of the ResNet-50 when tested against adversaries generated using the model's parameters from previous epochs. The accuracy drops correspond to epochs where stochastic gradient descent learning drops happen. Consistent with Figure \ref{fig:previous_adversaries}, we see that adversarial samples from the initial epochs are treated more or less like natural samples by the final model. The adversaries become more potent as the model parameters start to approach their final value and the model starts to stabilize. Samples from the initial phase of training have limited impact on improving robustness. In spite of this, the computationally expensive maximization \eqref{eq:pgd} is performed to generate these samples for training and these samples are allowed to influence the model parameters.

\begin{figure*}
    \centering
    \includegraphics[width=1\linewidth ]{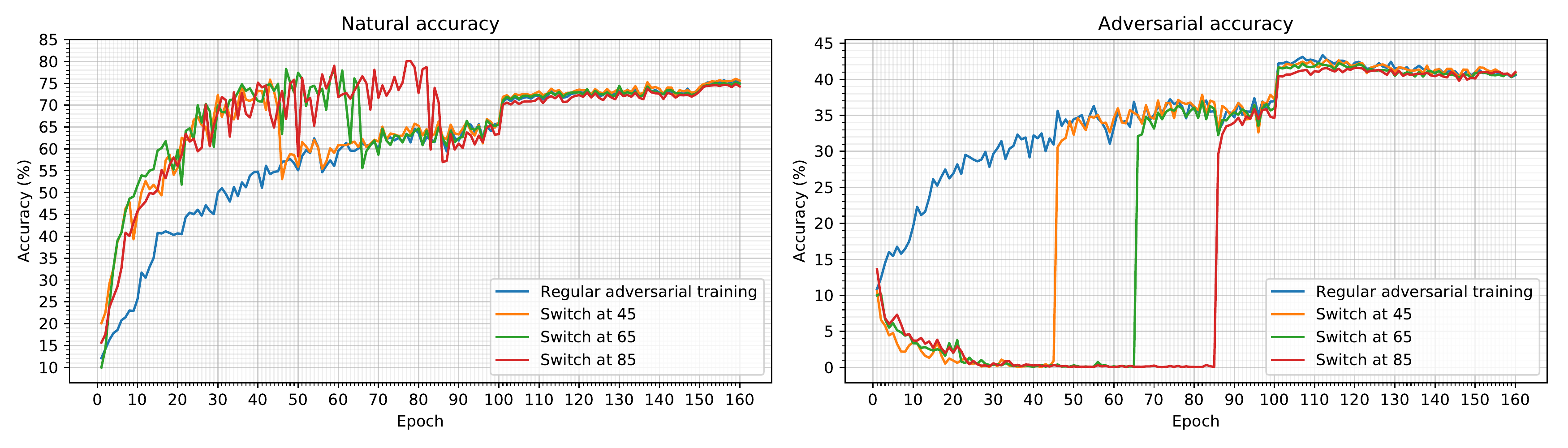}
    \caption{Natural and adversarial test accuracy during regular adversarial training and adversarial training with different switches. CIFAR-10 images are classified using the ResNet-50. Adversarial samples with $T = 10, \epsilon = \frac{8}{255}$ and $\alpha = \frac{2}{255}$ are used. SGD learning rate drops are after epochs 100 and 150. }
    \label{fig:resnet50_triggers}
\end{figure*}

\begin{figure*}
    \centering
    \includegraphics[width=1\linewidth ]{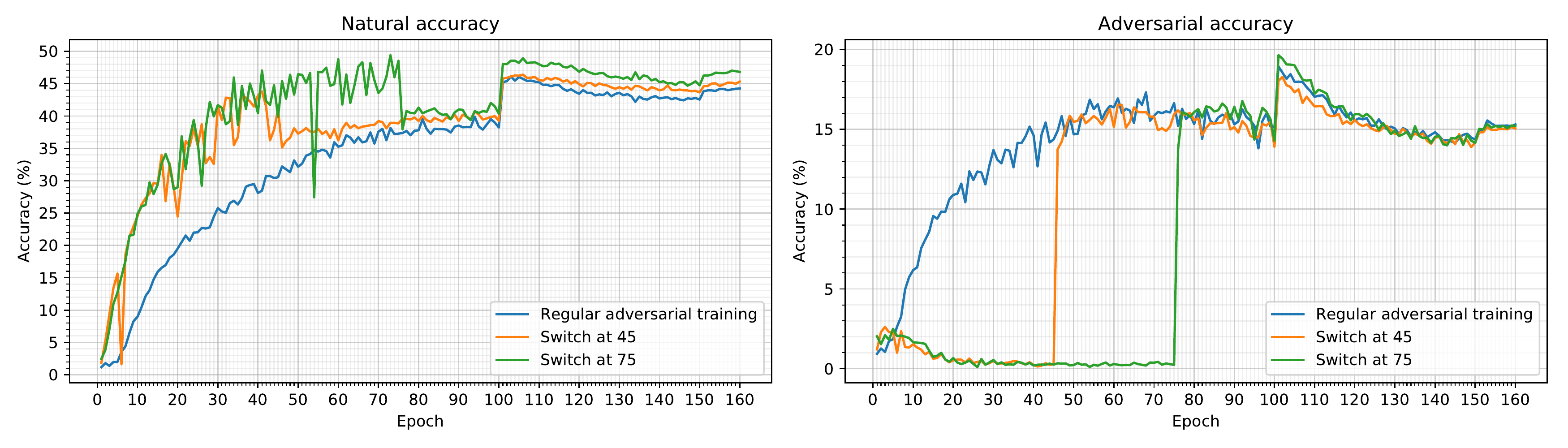}
    \caption{Natural and adversarial test accuracy during regular adversarial training and adversarial training with different switches. CIFAR-100 images are classified using the ResNet-50. Adversarial samples with $T = 10, \epsilon = \frac{8}{255}$ and $\alpha = \frac{2}{255}$ are used. SGD learning rate drops are after epochs 100 and 150. }
    \label{fig:resnet50_cf100_triggers}
\end{figure*}

Similar to Figure \ref{fig:wrn_triggers}, Figures \ref{fig:resnet50_triggers} and \ref{fig:resnet50_cf100_triggers} show the natural and adversarial accuracy during training when different switches are used. The plots are for CIFAR-10 and CIFAR-100 respectively.

\subsection{Additional ResNet-18 results} \label{sec:resnet18}

\begin{figure*}
    \centering
    \includegraphics[width=1\linewidth ]{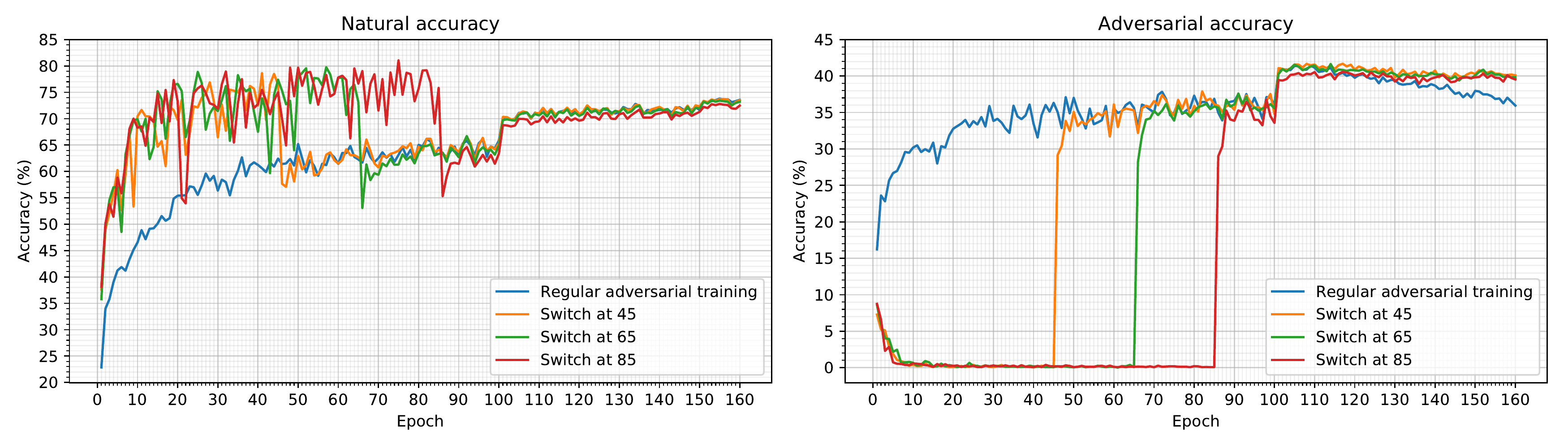}
    \caption{Natural and adversarial test accuracy during regular adversarial training and adversarial training with different switches. CIFAR-10 images are classified using the ResNet-18. Adversarial samples with $T = 10, \epsilon = \frac{8}{255}$ and $\alpha = \frac{2}{255}$ are used. SGD learning rate drops are after epochs 100 and 150. }
    \label{fig:resnet18_triggers}
\end{figure*}

\begin{figure*}
    \centering
    \includegraphics[width=1\linewidth ]{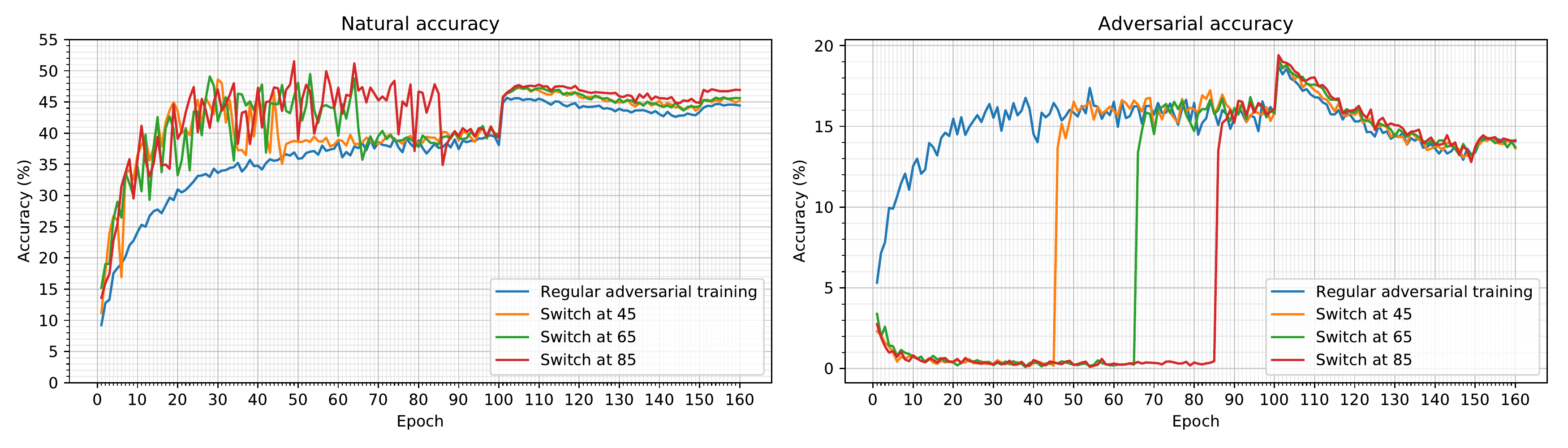}
    \caption{Natural and adversarial test accuracy during regular adversarial training and adversarial training with different switches. CIFAR-100 images are classified using the ResNet-18. Adversarial samples with $T = 10, \epsilon = \frac{8}{255}$ and $\alpha = \frac{2}{255}$ are used. SGD learning rate drops are after epochs 100 and 150. }
    \label{fig:resnet18_cf100_triggers}
\end{figure*}

Figures \ref{fig:resnet18_triggers} and \ref{fig:resnet18_cf100_triggers} show the natural and adversarial accuracy during training when different switches are used. Again we use the CIFAR-10 and CIFAR-100 datasets.

\begin{figure*}
    \centering
    \includegraphics[width=1\linewidth ]{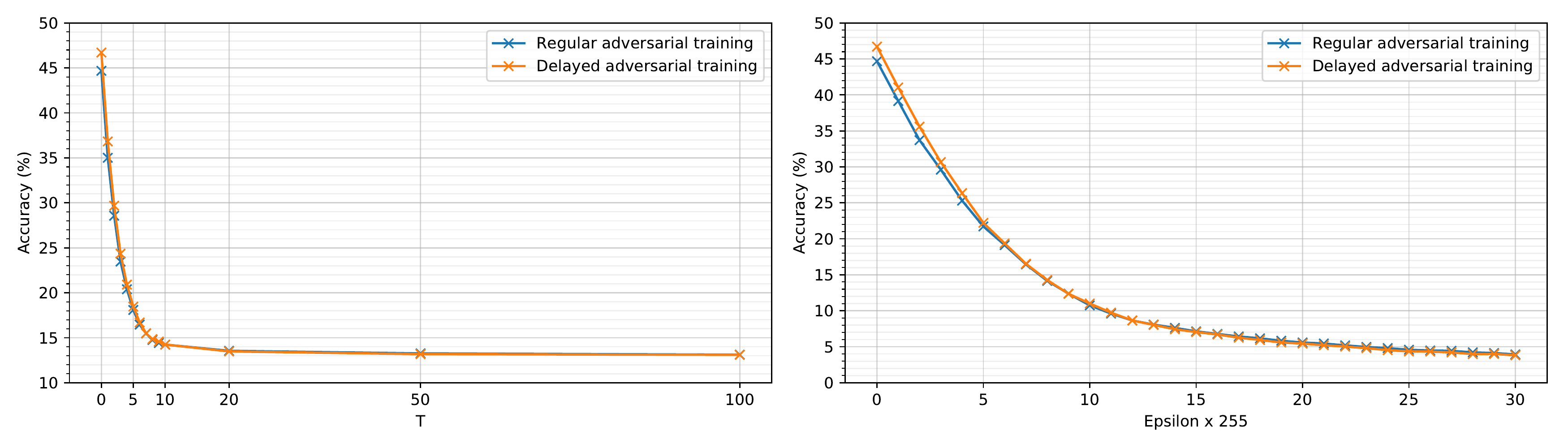}
    \caption{Accuracy of fully trained ResNet-18 with CIFAR-100 when tested with attacks of different strength. Adversaries used during training were of strength $\left\{10, \frac{8}{255}, \frac{2}{255}\right\}$.}
    \label{fig:resnet18_cf100_resistance}
\end{figure*}

Figure \ref{fig:resnet18_cf100_resistance} shows the robustness of ResNet-18 when trained on CIFAR-10 against adversaries which they were not trained to be robust against. Adversaries used during training were of strength $\left\{10, \frac{8}{255}, \frac{2}{255}\right\}$.

\subsection{Additional MNIST results}

\begin{figure*}
    \centering
    \includegraphics[width=1\linewidth]{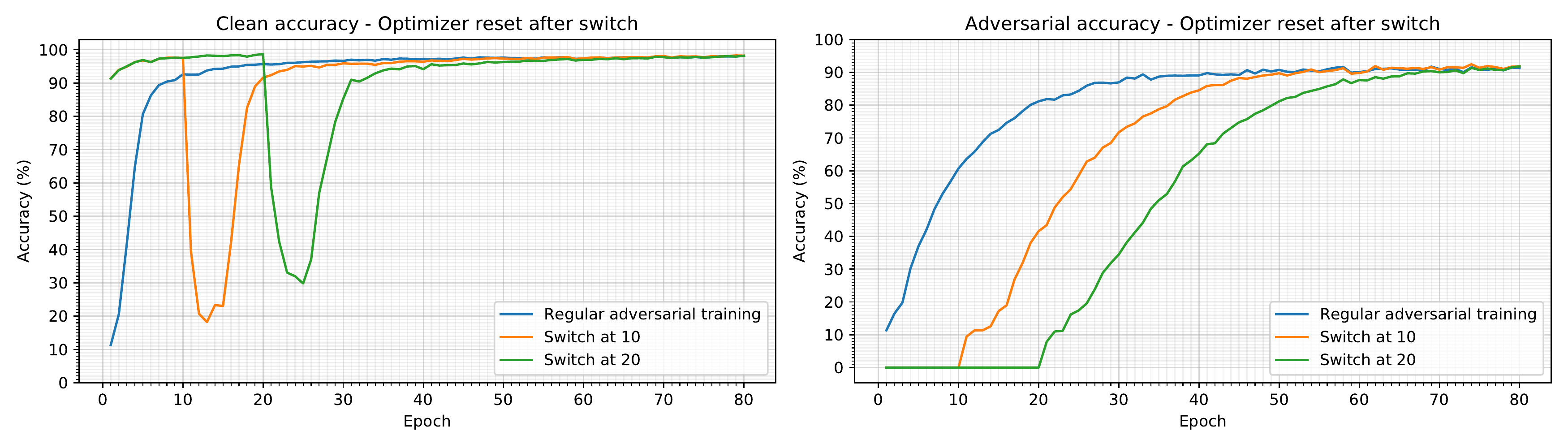}
    \caption{Natural and adversarial test accuracy during regular adversarial training and adversarial training with different switches. MNIST images are classified using two-layer CNNs. Adversarial samples with $T = 40, \epsilon = 0.3$ and $\alpha = 0.01$ are used.}
    \label{fig:mnist_accuracy_profiles_newopt}
\end{figure*}

Similar to Figure \ref{fig:wrn_triggers}, Figures \ref{fig:mnist_accuracy_profiles_newopt} shows the natural and adversarial accuracy during training when different switches are used. The plots are for MNIST with a two-layer CNN.

\section{Different seeds with WideResNet-28x10}

\begin{figure*}[t]
    \centering
    \includegraphics[width=1\linewidth ]{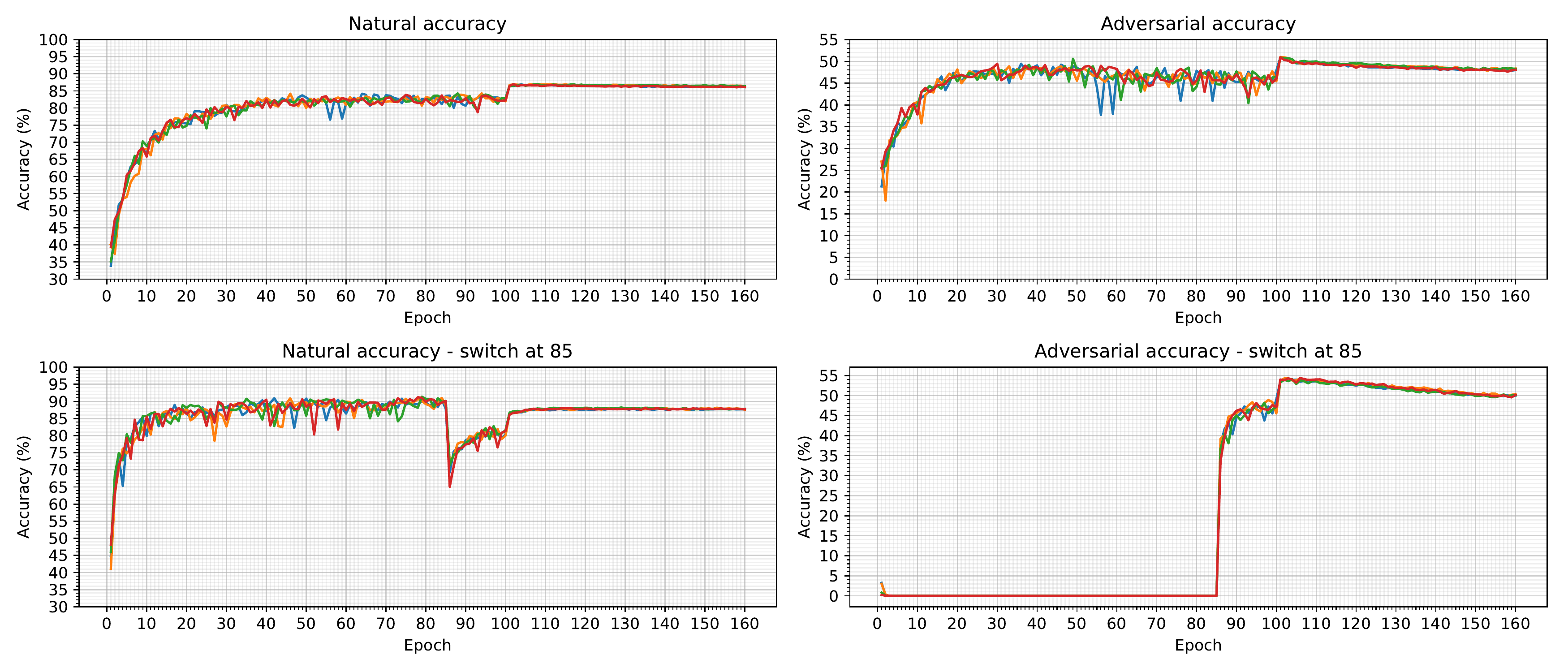}
    \caption{WideResNet-28x10 with CIFAR-10 initialized with different initial seeds. The natural and adversarial accuracy with and without switching is shown.}
    \label{fig:wrn_seeds}
\end{figure*}

Figure \ref{fig:wrn_seeds} shows that Delayed Adversarial Training works with different model parameter initialization seeds. The figure shows the WideResNet-28x10 being used for CIFAR-10 classification. The natural and test accuracy like in Figure \ref{fig:wrn_triggers} are plotted for different initialization seeds. The performance is similar across different seeds. Accuracy with and without switching is shown.

\end{document}